\documentclass[10pt,twocolumn,letterpaper]{article}

\usepackage{cvpr}      

\usepackage[dvipsnames]{xcolor}

\newcommand{\methodname}{VASE}
\newcommand{\supp}{\emph{Supp.~Mat.}}

\usepackage{pifont}
\newcommand{\cmark}{\ding{51}}%
\newcommand{\xmark}{\ding{55}}%

\newcommand{\myarrowup}{$(\uparrow)$}
\newcommand{\myarrowdown}{$(\downarrow)$}

\usepackage{colortbl}
\usepackage{mathtools}
\newcommand{\para}[1]{\noindent \textbf{#1.}}

\newcommand{\RNum}[1]{\uppercase\expandafter{\romannumeral #1\relax}}
\newcommand{\videoshape}[2]{\in \mathbb{R}^{#1 \times #2 \times H \times W}}
\newcommand{\videomaskshape}[2]{\in \{0,1\}^{#1 \times #2 \times H \times W}}

\newcommand{\imageshape}[1]{\in \mathbb{R}^{#1 \times H \times W}}
\newcommand{\imagemaskshape}[1]{\in \{0,1\}^{#1 \times H \times W}}

\newcommand{\sourcevideo}{\mathcal{X}}
\newcommand{\sourcemask}{\mathcal{M}}
\newcommand{\refimage}{I_{\text{ref}}}
\newcommand{\refshape}{M_{\text{ref}}}
\newcommand{\genvideo}{\mathcal{Y}}
\newcommand{\sourceflow}{\mathcal{F}}
\newcommand{\editregion}{M_{\text{edit}}}
\newcommand{\editset}{E}
\newcommand{\editopeartion}{\mathcal{E}}
\usepackage{multirow}
\usepackage{graphicx}
\usepackage[export]{adjustbox}

\definecolor{cvprblue}{rgb}{0.21,0.49,0.74}
\usepackage[pagebackref,breaklinks,colorlinks,citecolor=cvprblue]{hyperref}

\def\paperID{...} 
\def\confName{...}
\def\confYear{...}

\title{\methodname: Object-Centric Appearance and Shape Manipulation of Real Videos}

\author{Elia Peruzzo\,$^{\textcolor{cvprblue}{1}}$ \quad Vidit Goel\,$^{\textcolor{cvprblue}{2}}$ \quad Dejia Xu\,$^{\textcolor{cvprblue}{3}}$ \quad Xingqian Xu\,$^{\textcolor{cvprblue}{2}}$ \quad Yifan Jiang\,$^{\textcolor{cvprblue}{3}}$ \\ Zhangyang Wang\,$^{\textcolor{cvprblue}{2,3}}$ \quad Humphrey Shi\,$^{\textcolor{cvprblue}{2,4}}$ \quad Nicu Sebe\,$^{\textcolor{cvprblue}{1}}$ \\
\normalsize{$^{\textcolor{cvprblue}{1}}$ University of Trento \quad $^{\textcolor{cvprblue}{2}}$ Picasrt AI Research\quad $^{\textcolor{cvprblue}{3}}$ UT Austin \quad $^{\textcolor{cvprblue}{4}}$ Georgia Tech} 
}

\begin{document}
\twocolumn[{
    \renewcommand\twocolumn[1][]{#1}
    \maketitle
    \centering
\includegraphics{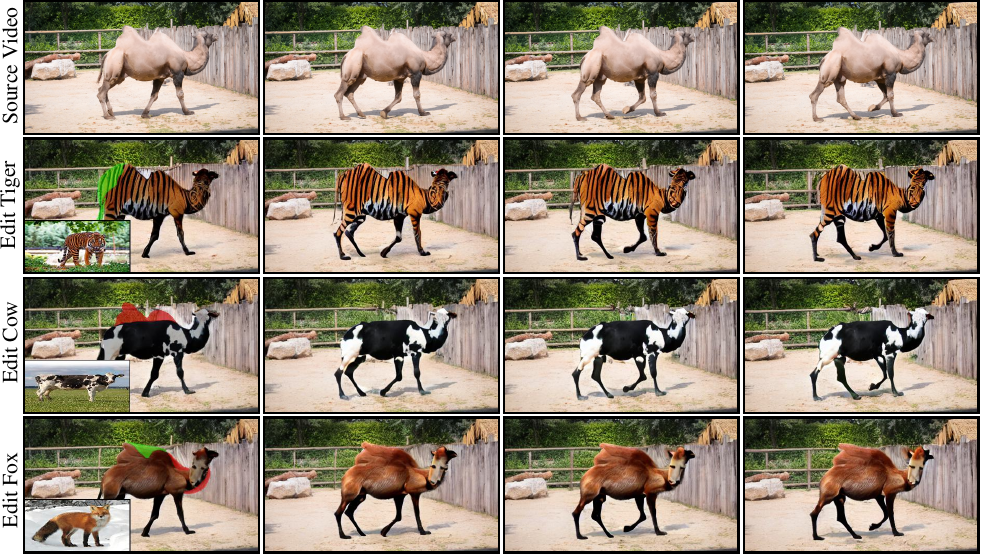}
\captionof{figure}{With {\methodname}, we can manipulate both the shape and appearance of an object within a real video. In the top row, we present the source video, followed by a series of edits showcasing precise adjustments in both attributes. In the first frame (1st column), the keyframe structure is overlayed to the object. \textcolor{Red}{Red} areas represent regions that have been removed from the original shape, while \textcolor{Green}{green} regions denote newly added parts. The driver image for appearance edits is showcased in the bottom left corner.}
\vspace{2em}
\label{fig:teaser}
}]

\begin{abstract}
Recently, several works tackled the video editing task fostered by the success of large-scale text-to-image generative models. However, most of these methods holistically edit the frame using the text, exploiting the prior given by foundation diffusion models and focusing on improving the temporal consistency across frames. In this work, we introduce a framework that is object-centric and is designed to control both the object's appearance and, notably, to execute precise and explicit structural modifications on the object.
We build our framework on a pre-trained image-conditioned diffusion model, integrate layers to handle the temporal dimension, and propose training strategies and architectural modifications to enable shape control. We evaluate our method on the image-driven video editing task showing similar performance to the state-of-the-art, and showcasing novel shape-editing capabilities. Further details, code and examples are available on our \href{https://helia95.github.io/vase-website/}{project page}.
\end{abstract}    
\section{Introduction}
\label{sec:intro}
In recent years, there has been a surge in both the quality and diversity of generated content, primarily due to the deployment of Diffusion Models (DM) \cite{sohl2015deep, song2019generative, ho2020denoising} trained on vast amounts of data \cite{schuhmann2021laion}. Remarkably, large-scale text-to-image (T2I) models \cite{rombach2022high, nichol2021glide, saharia2022photorealistic, ramesh2022hierarchical} enable inexperienced users to produce impressive results using only a textual prompt. 
A plethora of works have been developed ever since, tailoring these models for more specific and constrained use cases, enabling various editing capabilities and making them even more user-friendly \cite{chen2023control, Ruiz_2023_CVPR, palette2022saharia, huang2023composer}. Following a similar trend, many works are extending these paradigms to videos. The majority of current methods rely on a text prompt to guide the editing process \cite{wu2023tune, ceylan2023pix2video} and introduce new techniques to improve the smoothness and temporal consistency across the generated frames \cite{qi2023fatezero, yang2023rerender, geyer2023tokenflow}. As highlighted in prior research \cite{Ruiz_2023_CVPR, gal2022textual, goel2023pair}, text prompts, while intuitive, often fall short in capturing precise nuances, leading to potential mismatches with the user's intentions. Additionally, existing approaches treat the video frames as a whole for editing \cite{qi2023fatezero, geyer2023tokenflow, wu2023tune}, lacking the granularity needed to constrain the changes to a specific region. Lastly, the majority of these approaches cannot modify the structure of the objects in the video explicitly, as they often rely on per-frame structural guidance from the source video \cite{esser2023structure, chen2023control, wang2023videocomposer}, making them not suitable for this application.

In this work, our goal is to explore other directions in the video editing domain to empower the user with comprehensive capabilities. Specifically, we propose \textbf{\methodname}, a framework for video object-centric appearance and shape edits. Our approach is \emph{object-centric}, designed to control one individual object at a time while preserving the rest of the video intact. This mirrors numerous real-world scenarios where users seek to confine edits to a particular region. Secondly, we drive the edit using another image, as an image can convey more details than a textual prompt. More significantly, we introduce the potential to alter the object's structure, providing more comprehensive editing options. This capacity remains under-explored in previous methods, or if present, it lacks an explicit control mechanism \cite{lee2023shape, qi2023fatezero}. 
In this work, we argue that in a video editing setup, shape modifications should not alter the object's shape drastically, as this pertains more to a pure generative setup than an editing scenario. 

Our focus lies in enabling precise, user-driven shape adjustments by manipulating a \emph{single keyframe}, which then extends across the entire video. 
It is worth noting that our approach is not confined to particular domains (\eg humans); instead, we propose a generic framework that can modify both rigid and articulated objects for videos in the wild. Lastly, we exclude expensive per-video training or video-decomposition processes, which limit the practicality of editing methods. 

We base our method on a pre-trained image-conditioned diffusion model \cite{yang2023paint}, inflated with temporal layers to process videos. In order to faithfully replicate the motion depicted in the original video, we incorporate a ControlNet \cite{chen2023control} that takes the optical flow derived from the source video as input. Additionally, we unlock the explicit control over the object's shape by feeding the desired keyframe shape into the ControlNet.

In our initial implementation, we observed that the model could potentially solve the task by predominantly relying on the optical flow while disregarding the structural condition. This rendered the shape edits ineffective during inference. To address this, we propose a novel \emph{Joint Flow-Structure Augmentation} procedure to break the alignment between the optical flow and the shape conditioning. Additionally, we introduce an auxiliary \emph{Segmentation Head} at training time to force the model to rely on the structure information. Moreover,  we introduce a \emph{Flow-Completion Network} to generate realistic motion in the edited regions where no displacement information is available from the source video. 

To summarize, our key contributions are as follows:
\begin{itemize}
    \item We enable object-centric video editing in the wild, controlling the shape and the appearance of individual objects in the input video while keeping the rest of the video unchanged. The user provides one driving image that specifies the desired appearance and the shape of a single edited keyframe.
    \item We design a \emph{Joint Flow-Structure Augmentation} pipeline, and introduce a \emph{Flow-Completion Network} and a \emph{Auxiliary Segmentation Head}. All the components are necessary to edit the shape of the objects effectively.
    \item We perform challenging shape edits of the target object without relying on test-time optimization procedures. Our model is trained once and applied at inference time, enhancing its practicality for various editing scenarios.
\end{itemize}
\section{Related Works}
\label{sec:related_works}

\para{Image Editing with Diffusion Models}  
\looseness=-1
Large-scale text-to-image (T2I) diffusion models  \cite{rombach2022high, nichol2021glide, saharia2022photorealistic, ramesh2022hierarchical} have gained widespread adoption as powerful generative models. They serve as strong generative priors and can be tailored to accommodate various forms of conditioning, 
along with the ability to manipulate real images \cite{meng2021sdedit}.  One line of works, achieves editing capabilities in a zero-shot manner, by exploiting the internal representation of a pretrained T2I model. \citet{hertz2022prompt} investigate the role of cross-attention maps in the generated images and exploit them to bind edits to a specific region. 
Plug-and-Play \cite{tumanyan2023plug} instead processes a real image and a target text prompt, and exploits the features of self-attention blocks to preserve the structure from the original image while applying the editing operation. 
\citet{epstein2023diffusion} recently applies similar observation in conjunction with classifier-free guidance \cite{ho2022classifier} to achieve more comprehensive editing capabilities. 
Differently, ControlNet \cite{chen2023control} opts to train a hypernetwork to control the generation of the main T2I model based on spatial cues, enabling control with various types of signals \eg depth, semantic maps, or sketches. Concurrently, \cite{mou2023t2i} explores more lightweight designs for the hypernetwork and the combination of multiple conditioning signals. 
Other works explore image-driven conditioning as an alternative to text for specifying the desired appearance of the objects in the scene. In particular, personalization emerged as an important research direction \cite{Ruiz_2023_CVPR, gal2022textual, avrahami2023break}, where the goal is to inject a user-provided concept in the form of images which is not present in the training data. In a similar way, Paint-by-Example (PBE) \cite{yang2023paint} proposed the first image-based inpainting method with coarse masks depicting the shape of the desired object, and \cite{chen2023anydoor} enhances the fidelity to the driver image using features from strong self-supervised models. \cite{xu2023prompt} introduces a new encoder to capture more nuances of the driver image and implement a ``prompt-free" method, while PAIR Diffusion \cite{goel2023pair} extracts per-object appearance representation, which enables image-driven editing and precise control. We draw inspiration from these methods and aim for more comprehensive editing capabilities in the video domain.

\para{Video Generation and Editing with Diffusion Models}
Recently, there have been several efforts to expand the achievements of text-to-image models to the video domain \cite{singer2022make, ge2023preserve, ho2022imagen, Blattmann_2023_CVPR}. As foundational video diffusion models remain largely unavailable to the open-source community and prohibitive to train from scratch, many works explored available T2I models for video editing.

One line of work exploits pre-trained T2I models and adapts them to the task in a zero-shot manner \cite{qi2023fatezero, ceylan2023pix2video, khachatryan2023text2video, geyer2023tokenflow, yang2023rerender}. The temporal consistency of the generated frames is generally encouraged by inflating the self-attention blocks to process multiple frames \cite{khachatryan2023text2video, wu2023tune}. Tune-A-Video \cite{wu2023tune} involves fine-tuning the model on the video to be edited, enabling test-time edits through text prompts or cross-attention control \cite{liu2023video}. In FateZero \cite{qi2023fatezero}, the attention maps are extracted during an initial inversion step and blended with those generated during the editing process, confining the edit to a specific region. 
FateZero allows for shape modification of the foreground object, but it lacks a mechanism to specify the desired edit explicitly. TokenFlow \cite{geyer2023tokenflow} proposes to propagate features of the base T2I model leveraging the optical flow extracted from the source video. Differently, other methods opt for training on video datasets; these models employ factorized spatio-temporal layers, integrating pre-existing layers from a T2I model with newly initialized temporal blocks \cite{esser2023structure, wang2023videocomposer, guo2023animatediff}.
While these methods demonstrate remarkable editing capabilities with satisfactory temporal consistency, they do not explore the domain of shape manipulation.  Any potential changes in shape, if they occur, cannot be directly controlled, which poses a limitation on their applicability.

\para{Neural Layered Atlases (NLA)} \citet{kasten2021layered} propose to decompose the video in Neural Layered Atlases (NLA), exploiting the high correlation across frames. This representation is inherently designed for editing purposes, as the edits applied to the canonical image are propagated to the rest of the video, with temporal consistency obtained by design. Text2Live \cite{bar2022text2live} introduces text conditioning to obtain both image and video appearance editing results through text, while \cite{ouyang2023codef} improved the atlas decomposition both in terms of reconstruction quality and computational time. 
Recently, \citet{lee2023shape} leverages NLA representation in conjunction with an image editor and designed a framework that propagates shape edits. A deformation field is obtained with a semantic correspondence method between the input keyframe and the edited keyframe. Following that, the canonical representation of the foreground object is updated using the guidance of a T2I model through an optimization process \cite{poole2022dreamfusion}. Despite the remarkable results, the current version lacks direct control over the shape of the final object. Additionally, as pointed out by \cite{lee2023shape}, the pipeline is sensitive to the semantic correspondence method, which is inherently difficult to achieve with precision. Any discrepancies or inaccuracies in this process are reflected in the final video output. Furthermore, NLA-based methods have a high computational cost (approximately 10 hours for a 70-frame video) and may fail in case of complex motion and self-occlusions \cite{kasten2021layered}. Drawing inspiration from \cite{lee2023shape}, we aim to replicate similar capabilities in a model that doesn't necessitate NLA decomposition, making it a more practical alternative.
\section{Method}
\label{sec:method}

\begin{figure*}[!th]
    \centering
    \includegraphics[width=\linewidth, keepaspectratio]{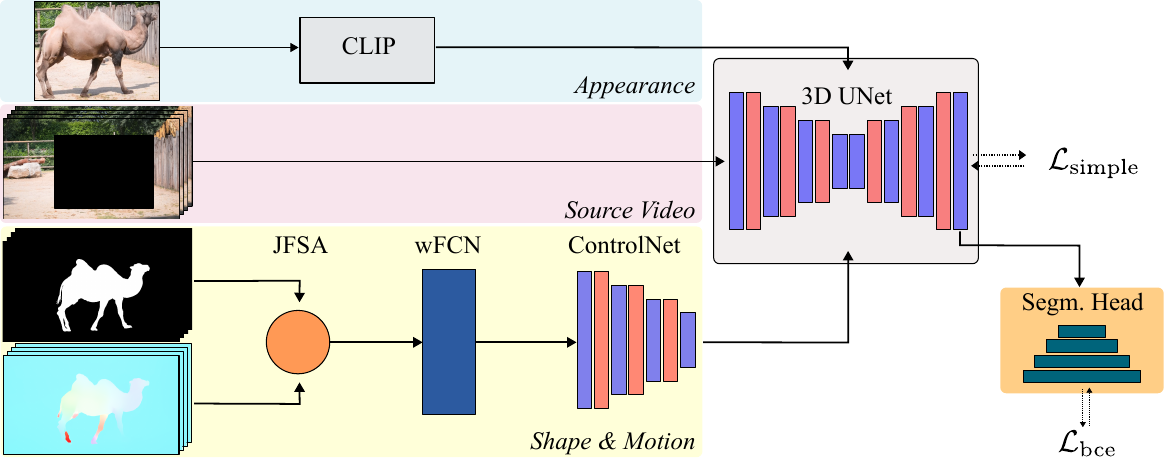}
    \caption{We enhance video editing by conditioning the synthesis of the video on two branches, one that controls the appearance and the other responsible for the motion and the structure of the object. To enable shape modifications, we propose a Joint Flow-Structure Augmentation pipeline that outputs an augmented flow which is processed by a Flow-Completion Network, before going as input to the final ControlNet module. Furthermore, we introduce an auxiliary loss, used to enhance the model fidelity to the input segmentation map.}
    \label{fig:method_figure}
\end{figure*}

{\methodname} is based on an image-conditioned video inpainting model. Our approach preserves the original video's motion by employing optical flow, and we introduce explicit guidance to enhance structural manipulations. We recall that the structure conditioning is provided for a single keyframe, which for simplicity we assume to be the first frame of the sequence. An overview of our pipeline is shown in \cref{fig:method_figure}.

Formally, let $\sourcevideo \videoshape{T}{3}$ be the source video and $\sourcemask\videomaskshape{T}{1}$ the segmentation mask of the foreground object targeted for editing. We denote with  $\sourcemask^{\text{bb}} \videomaskshape{T}{1}$ a bounding box mask enclosing it. Our goal is to synthesize an edited video $\genvideo \videoshape{T}{3}$, where the appearance of the foreground object is driven by $\refimage \imageshape{3}$, while the structure of the object is specified by $\refshape \imagemaskshape{1}$, which represents the shape of the edited keyframe. 
We define the edited region as $\editregion  \coloneqq \refshape \neq \sourcemask_0$, $\editregion \imagemaskshape{1}$.

\subsection{Backbone}
\looseness=-1
\label{sec:method_backbone}
\para{Inflated UNet} A significant challenge in video generation lies in achieving temporal consistency, which involves minimizing the flickering and inconsistencies between frames. The first step of our method is designed to tackle this problem. We follow recent works \cite{esser2023structure, Blattmann_2023_CVPR} and inflate a pretrained diffusion model by introducing temporal layers; we refer to this model as 3D Unet. Specifically, our temporal blocks are composed of a \emph{1D-convolution} followed by \emph{temporal self-attention}. The temporal blocks are placed after the original spatial blocks, and the input features are reshaped to accommodate the temporal dimension. 
We base our framework on Paint-by-Example (PBE) \cite{yang2023paint}, which serves as our base image-conditioned diffusion model. We adopt a similar self-supervised training strategy extended to videos and pre-train the model to reconstruct the source video $\sourcevideo$. The input to the model consists of the masked source video $\sourcevideo \odot ( 1-\sourcemask^{\text{bb}})$, and the reference object extracted from the $t$-th frame $\refimage \coloneqq \sourcevideo_t \odot \sourcemask^{\text{bb}}_t$. The model is trained with the $\mathcal{L}_{\text{simple}}$ loss \cite{rombach2022high}, defined as $\mathcal{L}_{\text{simple}} = \lVert \epsilon - \epsilon_\theta(z_t)\rVert_2^2$. In this first stage, the model learns to generate a temporally consistent sequence of frames.

\para{Inflated ControlNet} Next, we introduce a conditioning mechanism to guide the generation process, maintain the motion characteristics from the source video, and simultaneously add explicit control on the shape.
We extract the forward optical flow   $\sourceflow \videoshape{T-1}{2}$ from the source video using RAFT \cite{teed2020raft}. 
Then, we input to the network the structure of the keyframe $\refshape = {\sourcemask}_0$, \ie the segmentation mask of the first frame. 
We condition the model by incorporating a ControlNet \cite{zhang2023adding}, which takes $\{\mathcal{F}, \refshape\}$  as input, concatenated across the channel dimension. We pad the optical flow in the temporal dimension and repeat the structure conditioning $T$ times, to ensure compatibility of dimensions. During this stage, we train the ControlNet while freezing the Inflated UNet. 

This model serves as our backbone. While it can achieve appearance editing operations, it fails to perform shape editing control. During inference, if $\refshape \neq \sourcemask_0$, the model tends to overlook it. We impute this to the dominant signal coming from the optical flow $\sourceflow$, which can be used to minimize $\mathcal{L}_{\text{simple}}$ without considering the structure conditioning $\refshape$.

In the next subsections, we describe our proposed solutions to solve this problem and enable shape control at inference time.

\subsection{Joint Flow-Structure Augmentation (JFSA)}
\label{sec:method_jfsa}
As a first step, we propose an augmentation procedure to jointly modify the shape and the flow conditioning. As outlined in \cref{sec:intro}, we are focusing on edits of the foreground object, which can be reduced to two main types: adding a region or removing a region from the foreground.  With \emph{JFSA} we aim to simulate this behavior during training by modifying the optical flow $\sourceflow$ and the conditioning structural mask $\sourcemask$ in a systematic manner.
Specifically, we approximate regions of the target object that exhibit consistent motion across the $T$ frames by clustering the optical flow along the temporal dimension: 

\begin{equation} 
    \label{eq:clustering}
    \editset = \mathtt{KMeans}(\sourceflow + \lambda \cdot X_{\text{bias}}, \, N_c)
\end{equation}
Let $\editset = \{\editopeartion^0, \ldots, \editopeartion^{N_c} \}$ represent the set of clustered regions, where $\editopeartion^k \videomaskshape{T}{1}$ denotes the binary map associated with the $k$-th region. Here, $N_c$ stands for the number of clusters, $X_{\text{bias}}$ introduces a spatial bias, and $\lambda$ governs the trade-off.

We consider $E$ as a collection of possible ground truth shape-edits, and define two operations:

\begin{equation}
    \begin{split}
    \tilde{\sourceflow} & = \sourceflow; \\
    \tilde{\sourcemask} & = \sourcemask \odot (1 - \editopeartion^k)
    \end{split}
    \label{eq:mask_shape}
\end{equation}
and,
\begin{equation}
    \begin{split}
        \tilde{\sourceflow} & = \sourceflow \odot (1 - \editopeartion^k) + \Big( \dfrac{1}{H \cdot W} \sum_{i,j} \sourceflow(i,j) \Big) \cdot \editopeartion^k; \\
        \tilde{\sourcemask} & = \sourcemask
    \end{split}
    \label{eq:mask_flow}
\end{equation}

Respectively, in \cref{eq:mask_shape}, we address the scenario where the user intends to remove a region from the foreground object. Meanwhile, in \cref{eq:mask_flow}, we simulate the scenario where the user aims to include an additional region in the foreground object.
This procedure is efficient and can be executed on the fly during training. We randomly choose an edit region $\editopeartion^k$ and perform one of the two operations with a specified probability $p_{\text{augm}}$. Note that we use $\tilde{\sourceflow},\tilde{\sourcemask}$ as the updated signals, replacing the original $\sourceflow, \sourcemask$, and use them as described in \cref{sec:method_backbone}.

\subsection{Warping Flow-Completion Net (wFCN)}
\label{sec:methdod_wfcn}
Inspired by the Video Inpainting literature \cite{zhou2023propainter}, we introduce an additional network that explicitly predicts a \emph{complete} optical flow before passing it to the ControlNet. 
The objective of this network is to estimate the optical flow within the editing area, thus simplifying the task for the subsequent synthesis module. Unlike Video Inpainting methods, our approach involves using a single mask for the initial keyframe instead of employing individual masks for each frame in the video. To bridge this gap, we propose to employ forward warping \cite{niklaus2020softmax} as an initial estimation for the edited region's location. It is important to note that although forward warping may introduce holes or inconsistencies, in our case, it is applied to a binary mask, mitigating these issues. 
We repeatedly warp the edited region obtaining $\tilde{\editopeartion}^k = W_{\text{sum}}(\editopeartion^k_0, \sourceflow)$, $\tilde{\editopeartion}^k \videomaskshape{T}{1}$ , and $W_{\text{sum}}$ the summation splatting operation as defined in \cite{niklaus2020softmax}.
Note that, while at training time the ground truth temporal edit operation $\editopeartion^k$ is available, we obtain better results by warping the mask instead, to reduce the gap with inference. Conversely, during the inference phase, the source optical flow $\sourceflow$ may be unavailable, particularly when adding a region to the foreground object. In such cases, infilling the missing region using the displacement of the nearest known point belonging to the object provides a reliable approximation, effectively resolving the issue (see {\supp} for details).
Our Warping Flow-Completion Network (wFCN) takes as input the concatenation of the optical flow, the warped edited region, and the structural mask, and predicts the complete optical flow $\sourceflow_{\text{complete}} = \text{wFCN}(\tilde{\editopeartion}^k, \sourceflow, \tilde{\sourcemask})$.
We pre-train this network offline using our \emph{JFSA} procedure to obtain pairs of editing regions and optical flow. During this phase, the primary driving loss is the reconstruction loss $\mathcal{L}_{\text{MSE}}(\sourceflow_{\text{complete}}, \tilde{\sourceflow})$, along with the other regularization losses defined in \cite{zhou2023propainter}.
The complete optical flow $\sourceflow_{\text{complete}}$ is utilized in conjunction with $\tilde{\sourcemask}$ as input for the ControlNet.

\subsection{Segmentation Head (SH)}
\label{sec:method_sh}
To emphasize the importance of shape conditioning, we incorporate a lightweight segmentation head during the training process. More specifically, we extract the second-to-last feature of the 3D UNet and pass it through a small convolutional decoder, which receives as additional input the embedding of the diffusion timestep. We supervise the model with binary cross-entropy loss $\mathcal{L}_{\text{bce}}(\sourcemask_{\text{pred}}, \tilde{\sourcemask})$, with $\sourcemask_{\text{pred}}$ the predicted segmentation maps. We optimize it together with the main loss of the diffusion model; the total loss thus becomes $\mathcal{L}_{\text{tot}} = \mathcal{L}_{\text{simple}} + \alpha \cdot \mathcal{L}_{\text{bce}}$.
We further exploit the segmentation predictions $\sourcemask_{\text{pred}}$ at inference time, to preserve the background within the unpainted area, performing masked DDIM sampling \cite{rombach2022high}. We refer to the {\supp} for additional details.

\begin{table*}[!hbt]
    \centering
    \begin{tabular}{lccccccc}
        \toprule
        \multirow{2}{*}{\textbf{Method}}& \multicolumn{2}{c|}{\textbf{Video Quality}} & \multicolumn{2}{c|}{\textbf{Temporal Consistency}} & \multicolumn{3}
        {c}{\textbf{Image Alignment}} \\ \cmidrule{2-8}
        \multicolumn{1}{c}{} & DOVER \myarrowup & \multicolumn{1}{c|}{FVD\myarrowdown} & WE\myarrowdown & \multicolumn{1}{c|}{CLIP-T \myarrowup} & CLIP-S \myarrowup & DINO-S \myarrowup & LPIPS \myarrowdown \\ 
        \midrule

        PAIR Diffusion \cite{goel2023pair} & 0.602 &  24.32 & 19.6 & 0.904 & 0.552 & \textbf{0.36} & \textbf{0.71 }\\
        FILM \cite{reda2022film} & 0.643 & 19.85 & 8.9 & 0.956 & 0.546 & 0.33 & 0.76 \\
        TokenFlow \cite{geyer2023tokenflow} & \textbf{0.675} & 18.67 & \textbf{6.1} & \textbf{0.957}  & 0.501 & 0.29 & 0.81 \\ 
        \rowcolor{gray!25} \methodname & 0.672 & \textbf{16.54} & 9.8 &  0.955 & \textbf{0.567} & \textbf{0.36} & 0.73  \\
        \bottomrule
    \end{tabular}
    \caption{Quantitative results for the task of Image Driven Appearance Editing. FVD and WE have been rescaled by a factor of $10^2$, $10^{-3}$.}
    \label{tab:metrics_appearnace_editing}
\end{table*}
\begin{figure*}
    \centering
    \includegraphics{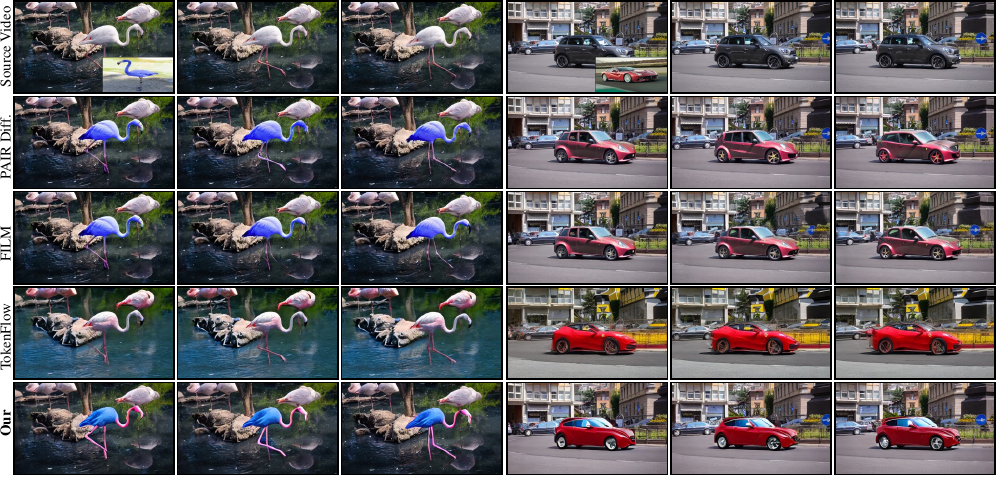}
    \caption{Results for the Image Driven Appearance Editing task. The driver image is displayed in the bottom-left corner of the initial frame in the source video. We refer the reader to the \textbf{{\supp}} for video results that better showcase the differences.}
    \label{fig:apperance_editing}
    \vspace{-2.5mm}
\end{figure*}

\subsection{Implementation Details}
Our foundation model is Paint-by-Example (PBE) \cite{yang2023paint}, a latent diffusion model \cite{rombach2022high} trained for the task of image-based inpainting. We inherit the appearance conditioning mechanism from PBE, \ie the driver image is embedded through CLIP \cite{radford2021learning}. We add temporal layers to the network in the form of 1d convolution and temporal self-attention, which are initialized following previous work \cite{Blattmann_2023_CVPR}. The architecture of the \emph{Flow Completion Network} is borrowed from ProPainter \cite{zhou2023propainter}, and adapted to accommodate one additional channel as input to process the segmentation map. Lastly, the \emph{Segmentation Head} is implemented as 4 convolutional layers, which upsample the second-to-last feature of the 3D Unet decoder (\ie before the final \texttt{conv-out} layer) by a factor of $8$. Our training procedure can be divided into three stages: initially, we train the 3D UNet on the video dataset and keep it frozen in the next steps. Subsequently, we introduce the ControlNet, conditioning on the original optical flow and segmentation map from the source video.
Finally, we integrate the pretrained \emph{wFCN} and the \emph{Segmentation Head} into the model and jointly train them along with the ControlNet using our \emph{JFSA} procedure. 

We use  YouTube-VOS \cite{xu2018youtube} for training, and resize the frames at a resolution of  $448\times256$. We train the model for a total of $400K$ iterations. During the inference stage, we employ DDIM sampling with 50 steps to generate the edited video. Additionally, we integrate classifier-free guidance \cite{ho2022classifier} to enhance the quality of the final output. We refer to the {\supp} for more comprehensive implementation details. 
\section{Experiments}
\label{sec:exp}

We showcase the performances of {\methodname} both qualitatively and quantitatively. 
We study two scenarios: \emph{Image-driven Appearance Editing}, wherein the structure of the source video remains unchanged, and the edit is driven by a reference image, and \emph{Joint Appearance-Shape Editing}, where both the object's structure and appearance are edited simultaneously.
\begin{table*}[!hbt]
    \centering
    \begin{tabular}{lccccc}
        \toprule
        \multirow{2}{*}{\textbf{Method}}& \multicolumn{1}{c|}{\textbf{Video Quality}} & \multicolumn{2}{c|}{\textbf{Temporal Consistency}} & \multicolumn{1}
        {c|}{\textbf{Shape Guidance}} & \multicolumn{1}{c}{\textbf{Throughput}} \\ \cmidrule{2-6}
        \multicolumn{1}{c}{} & \multicolumn{1}{c|}{DOVER \myarrowup}  & WE \myarrowdown & \multicolumn{1}{c|}{CLIP-T \myarrowup} & \multicolumn{1}{c|}{mIoU \myarrowup} & Hz \myarrowup \\ \midrule
        Shape-NLA \cite{lee2023shape} & 0.543 &  \textbf{5.2} & \textbf{0.970} & 0.83 & 0.11 \\
        \rowcolor{gray!25} \methodname & \textbf{0.565}  & 10.2 & 0.962 & \textbf{0.89} & \textbf{0.35} \\ 
        \bottomrule
    \end{tabular}
    \caption{Quantitative comparison with Shape-NLA \cite{lee2023shape} on the Joint Appearance-Shape Editing task. WE has been rescaled by $10^{-3}$.}
    \label{tab:metrics_shape_editing}
\end{table*}
\begin{figure*}
    \centering
    \includegraphics{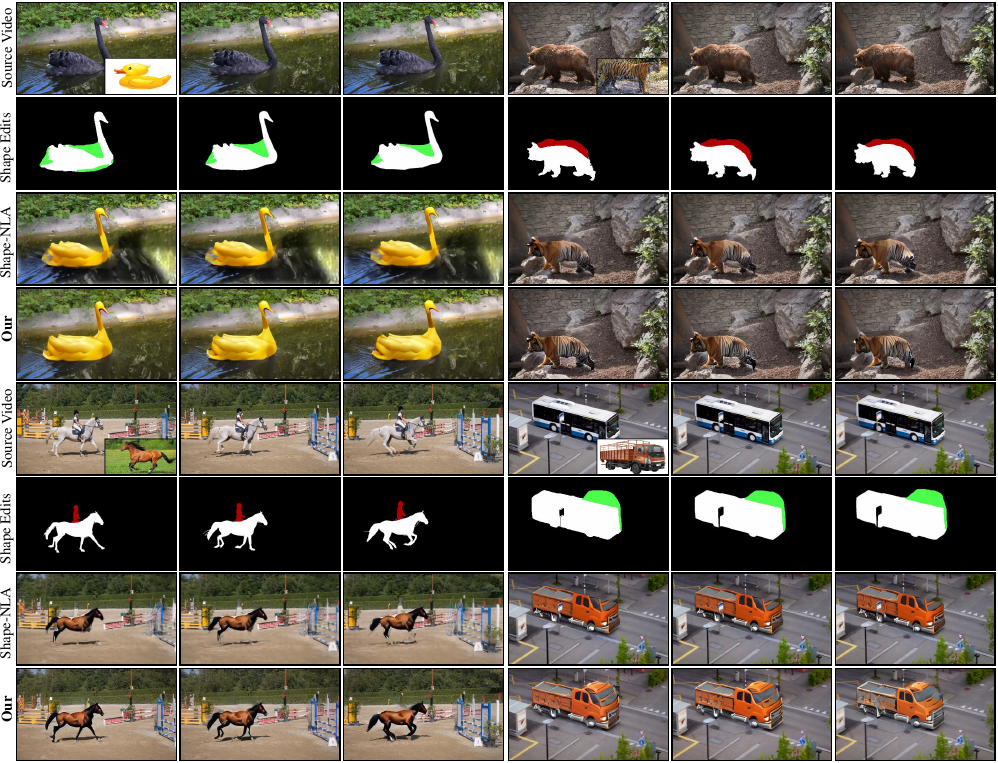}
    \caption{Results for the Joint Appearance-Shape Editing task. Please, note that the shape edit is provided \textbf{only} for the first keyframe, we show all of them only for visualization purposes. We highlight the structure modifications either in \textcolor{Red}{red} or \textcolor{Green}{green}.}
    \label{fig:shape_editing}
    \vspace{-2mm}
\end{figure*}
\subsection{Image-Driven Appearance Editing}
\label{sec:exp_image_driven}
The model is tasked with editing a target object in a source video based on the guidance from a driver image. It is important to emphasize that both the structure of the object and the background should remain unchanged.

\para{Baselines} We compare our method with the following baselines. (i) \emph{Frame Independent Editing}: we employ PAIR Diffusion \cite{goel2023pair} to independently edit each frame of the source video. 
(ii) \emph{Multi Frame Editing}: we use FILM \cite{reda2022film} to perform temporal interpolation between frames edited using the frame independent baseline \cite{goel2023pair}. A stride of 4 is utilized in this process, obtaining videos with superior temporal consistency. (iii) \emph{Text-based Editing}: we compare with TokenFlow \cite{geyer2023tokenflow}, and use BLIP2 \cite{li2023blip2} to caption the driver image before editing.

\para{Metrics} We evaluate the competitors on different aspects \cite{liu2023evalcrafter}: (i) \emph{Video Quality}: we rely on Fr\'echet Video Distance (FVD) \cite{unterthiner2018towards, unterthiner2019fvd} to compute a score for the overall video quality. Moreover, we follow recent work and compute a per-video quality score using DOVER \cite{wu2023dover}. (ii) \emph{Temporal Consistency}: we compute the Warping Error (WE) \cite{lai2018learning}, as the pixel-wise difference between the warped next frame and the current frame. Additionally, we measure the CLIP similarity between adjacent frames (CLIP-T) \cite{esser2023structure, geyer2023tokenflow}. Given our focus on editing a particular foreground object, our computations are confined to the region that encloses it, without considering the background. (iii) \emph{Image Alignment}: we evaluate the faithfulness of the edited frame to the driver image; we use the cosine similarity based on CLIP \cite{radford2021learning} (CLIP-S) and DINO \cite{oquab2023dinov2}  (DINO-S) features, and use the LPIPS  to capture more low-level details \cite{zhang2018unreasonable}. 

\para{Dataset} We create a benchmark, using 30 videos from the DAVIS dataset \cite{Perazzi2016} and selecting 3 driver images to edit each video. All the methods are evaluated at a resolution of $448 \times 256$, with $24$ fps.

\para{Results} We report the quantitative results in \cref{tab:metrics_appearnace_editing}. We can observe that {\methodname} archives good results in terms of overall visual quality. Furthermore, it is generally more aligned with the driver image, missing some low-level details compared to \cite{goel2023pair}. Lastly, while {\methodname} shows slightly lower numbers in terms of temporal consistency, we remark that it is not the main focus of the paper, and better consistency is expected by training on a higher quality training set both in terms of fps and visual quality (\eg WebVid \cite{bain2021frozen}).
Analyzing the qualitative results in \cref{fig:apperance_editing}, we can observe how the text-level conditioning of TokenFlow sometimes misses important aspects of the driver image (\eg the color of the flamingo), and leaks the edit to the background (second column). Conversely, results obtained with PAIR Diffusion exhibit strong flickering artifacts, whereas FILM enhances the temporal consistency at the expense of introducing distortions in the videos. For a more comprehensive video comparison, we direct the reader to the {\supp}.

\subsection{Joint Appearance-Shape Editing}
\label{sec:exp_shape_editing}
The objective is to alter the structure of one object in a source video, guided by the initial edited keyframe; simultaneously, a driver image is employed to edit the appearance. Note that the edit is restricted to one specific object,  and the background should remain unaffected. 

\para{Baselines} We compare our method with \cite{lee2023shape}, and denote it with Shape-NLA for brevity. Specifically, we edit the key-frame with \cite{goel2023pair}. Next, we construct the dense semantic mapping, by assigning each point in the edited area to either the closest point in the background or the foreground.

\para{Metrics} Following the previous setting, we compare the two methods on the \emph{Video Quality} and \emph{Temporal Consistency}. We use the metrics defined in \cref{sec:exp_image_driven}, excluding FVD, as it could not be reliably computed on a limited set of samples. Furthermore, we analyze two other aspects. (i) \emph{Shape Control}: to assess if the shape edit is reflected in the final frame. Evaluating it for the entire sequence presents challenges, primarily due to the absence of ground truths for every frame. Consequently, our assessment focuses solely on the first frame, calculating the mIoU. (ii) \emph{Throughput}: we quantify the inference cost of the different methods, by reporting the number of frames processed per second (Hz). It is noteworthy that, for Shape-NLA \cite{lee2023shape}, the NLA decomposition step is \underline{excluded} from this computation. This step demands roughly an additional 10 hours per video \cite{kasten2021layered}.

\para{Dataset} We use a subset of our benchmark used in \cref{sec:exp_image_driven} composed of 10 videos. For each video, we manually craft a structural edit for the initial frame.

\para{Results} We report the quantitative results on \cref{tab:metrics_shape_editing}. We can observe that Shape-NLA \cite{lee2023shape}, obtain excellent performances in term of temporal consistency thanks to the NLA decomposition. However, the faithfulness to the edited shape is inferior compared to our method, as shown by a lower mIoU. We can observe it in the qualitative results displayed in \cref{fig:shape_editing}. Shape-NLA struggles to precisely edit the shape \eg in the \texttt{blackswan} and \texttt{bus} examples. At the same time, it produces artifacts around the edited shape, \eg \texttt{tiger}. 
Lastly, the NLA decomposition introduces errors in the foreground object (for instance the legs in the \texttt{horsejump-low} case) and distortion in the background.

\subsection{Ablation}
\begin{table}[t]
    \centering
    \resizebox{\columnwidth}{!}{
    \begin{tabular}{lccccccc}
        \toprule
        \multicolumn{1}{l|}{\textbf{Version}} &
        \textbf{JFSA} & \textbf{wFCN} & \multicolumn{1}{c|}{\textbf{SH}} & \multicolumn{1}{c|}{\textbf{DOVER}} & \textbf{WE} & \multicolumn{1}{c|}{\textbf{CLIP-T}} & \textbf{mIoU} \\
        \midrule
        \RNum{1} &   \xmark & \xmark & \xmark & 0.514 & 13.2 & 0.939 & 0.79 \\
        \RNum{2} & \cmark & \xmark & \xmark & 0.529 & 12.2 & 0.945 & 0.84 \\
        \RNum{3} & \cmark & \cmark & \xmark & 0.546 & 11.7 & 0.952 & 0.87 \\
        \rowcolor{gray!25}Full & \cmark & \cmark & \cmark & \textbf{0.565} & \textbf{10.2} &  \textbf{0.962} &  \textbf{0.89} \\
        \bottomrule
    \end{tabular}}
    \caption{Ablation study on the components of \methodname.}
    \label{tab:arch_ablation}
\end{table}

\begin{figure}[!t]
    \centering
    \includegraphics{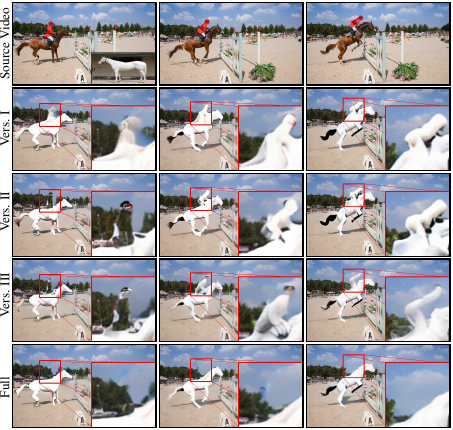}
    \caption{Qualitative results obtained with different versions of our model.}
    \label{fig:ablation_figure}
    \vspace{-2mm}
\end{figure}
We showcase results obtained with different variations of our model in \cref{fig:ablation_figure}, and compare them quantitatively in \cref{tab:arch_ablation}. We employ the experimental setup outlined in \cref{sec:exp_shape_editing} for the comparison. Specifically, we train three versions of our full model. In the first experiments, we keep only the backbone model as described in \cref{sec:method_backbone} \textbf{(Vers. I)}. We can notice that the shape modification is ignored by the model, which focuses only on the optical flow to solve the task.
Secondly, we add the Joint Flow-Structure Augmentation procedure \cref{sec:method_jfsa} (\textbf{Vers. II}). In this case, the model produces the desired edit in the first frame but generates artifacts around the object and the edit vanishes in the subsequent frames. Next, we introduce the Flow Completion Network \cref{sec:method_sh} \textbf{(Vers. III)}, which results in edits that are propagated to the whole sequence but with severe artifacts around the borders. Lastly, incorporating the Segmentation Head, we show the result of our \textbf{Full} model, which is the only one achieving satisfactory results.
We refer the reader to the {\supp} for further analysis and details.

\section{Conclusions}
\label{sec:conclusions}
In this paper, we present {\methodname}, a framework that enables users with a high degree of control over real video editing. We propose an object-centric formulation, that enables appearance modifications through a driver image and precise structural modifications with a single keyframe. While our method is able to produce a variety of satisfactory edits, it presents limitations in the presence of strong occlusions or significant changes in perspective for the target object. Furthermore, achieving consistent edits in lengthy videos remains a challenge. We identify these limitations as opportunities for future exploration, possibly incorporating 3D information to make the framework more robust.

{
    \small
    \bibliographystyle{ieeenat_fullname}
    \bibliography{main}
}

\clearpage
\setcounter{page}{1}
\maketitlesupplementary
\appendix

We present additional details and results for our method. 
Specifically, in \cref{sec:supp_method}, we delve into a more comprehensive discussion of the components of {\methodname}. We provide visualization for the \emph{Joint Flow-Structure Augmentation} procedure, along with the predictions of the \emph{Segmentation Head} and the \emph{Warping Flow Completion Network}. In \cref{sec:supp_implementation} we collect the implementation details of our method. Lastly, in \cref{sec:supp_qualitatives} we provide additional qualitative results of our model, both for the image-driven appearance editing and for the join structure and appearance editing task. We refer the reader to the \href{https://helia95.github.io/vase-website/}{project page} for full video results.
 
\section{VASE components}
\label{sec:supp_method}
In this section, we provide details about the main components of {\methodname} and show their respective effect with qualitative results. For a quantitative comparison and an examination of the specific contribution of each component, we refer the reader to \cref{tab:arch_ablation} of the main paper.

\begin{figure}[tb]
    \centering
    \includegraphics{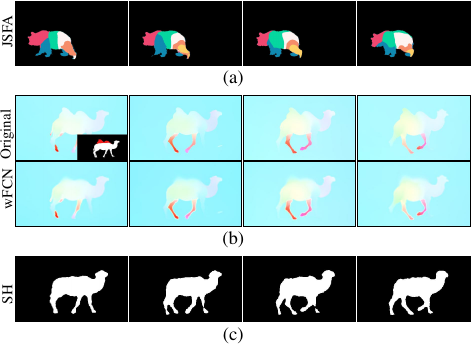}
    \caption{Qualitative results showcasing components of {\methodname}. (a) Illustrates JSFA, where distinct colors signify different clustered regions.  (b) Compares the original optical flow extracted from the source video (top row), with the complete optical flow predicted by the wFCN (bottom row). The edit region is shown at the bottom of the original flow. (c) Shows the corresponding predicted segmentation map.}
    \label{fig:supp_jsfa}
\end{figure}

\para{Joint Flow-Structure Augmentation} This augmentation procedure simulates structural modifications of the target object during training. To achieve realistic edits, our approach involves clustering the optical flow along the temporal dimension. This results in regions that exhibit consistent motion within the $T$ frames. As outlined in \cref{eq:clustering}, we introduce a regularization term in the form of a spatial bias, to promote the assignment of spatially neighbor pixels to the same region. The bias is computed as
\[
X_{\text{bias}}(i,j) = (i, j) \quad \forall i \in \{0, \ldots, H-1\}, \, \forall j \in \{0, \ldots, W-1\}
\]

and replicated $T-1$ times before adding it to the optical flow $\sourceflow$. We employ KMeans clustering and vary the number of cluster centroids $N_c$ within a random range of 6 to 10. In \cref{fig:supp_jsfa} (a), we present a qualitative example of the JFSA procedure. Here, distinct colors signify different regions clustered by our approach.

\para{Warping Flow Completion Network} We present details about the warping operation conducted before the Flow Completion Network. This operation is introduced to align with the original use case of Video Inpainting \cite{zhou2023propainter}, where a mask is supplied for each frame. Additionally, while not entirely precise, this approach facilitates the task for the network. We use summation splitting \cite{niklaus2020softmax} prior to feeding the inputs to the flow completion network. During inference, two cases arise. If a region is removed from the object, the original optical flow is readily available. On the other hand, if we add a region to the object, the flow is absent, leading to potential warping errors. To address this issue, we employ a displacement determined by the closest point in the object for which the flow is available. The nearest neighborhood search is based on the Euclidean distance between the points. This provides us with an approximation for the missing flow region, enabling us to warp the edit region and facilitating the subsequent task for the wFCN. 

We showcase an example of the complete optical flow predicted by the wFCN in \cref{fig:supp_jsfa} (b).

\para{Segmentation Head} We present qualitative results of the segmentation head's predictions during inference in \cref{fig:supp_jsfa} (c). These masks are utilized for masked DDIM sampling, with further details provided in the subsequent subsection.

\section{Model and Implementation Details}
\label{sec:supp_implementation}
\para{Architecture} Our foundation model is Paint-by-Example (PBE) \cite{yang2023paint}, a latent  diffusion model \cite{rombach2022high} trained for the task of image-based inpaining. We inherit the appearance conditioning mechanism from PBE, \ie the driver image is embedded through CLIP, and the obtained representation is further compressed with a linear layer which reduces its dimension, to avoid copy-paste artifacts. Next, we inflate the UNet architecture by adding temporal layers between the original spatial layers. Specifically, the original spatial layers of PBE process each frame independently, producing features with dimensions $f_{\text{spatial}} \in \mathbb{R}^{(\text{bs}\cdot\text{T})\times\text{dim}\times\text{h}\times\text{w}}$. The features are reshaped to expose the temporal dimension before being forwarded to the temporal layers, \ie $f_{\text{temporal}} \in \mathbb{R}^{(\text{bs}\cdot\text{h}\cdot\text{w})\times\text{dim}\times\text{T}}$. Finally, they are reshaped back to the original spatial structure. The temporal layers consist of a 1D-Convolution with a kernel size of 3, followed by a self-attention block, with relative positional encoding to capture the temporal order. The ControlNet architecture mirrors the encoder branch of the UNet, including the newly added temporal layers. We use a singular ControlNet, which receives the optical flow and the segmentation maps concatenated across the channel dimension.

The \emph{Flow Completion Network} architecture is borrowed from ProPainter \cite{zhou2023propainter}. We adapt the architecture and add one channel to the input layer, which receives a concatenation of the source optical flow, the warped edited region, and the segmentation mask.

The last component of our framework is the Segmentation Head, which is implemented as 4 convolutional layers. It upsamples the second-to-last feature of the 3D Unet decoder by a factor of $8\times$ and produces an output that matches the dimensions of the input frames before the VQGAN encoding \cite{rombach2022high}. This head is trained using binary cross-entropy (bce) loss, with a weighting factor set to $\lambda=0.05$.

\para{Training} Our training procedure can be divided into three stages. (i) Train the 3D Unet on a video dataset, in this stage the network learns to model the temporal dimension and the consistency between adjacent frames. This stage is performed for 250K iterations. (ii) Introduce the ControlNet with flow conditioning and shape control while freezing the 3D Unet, in this way, the network learns to preserve the motion from the optical flow. In this stage, the model is trained for 100K iterations. (iii) Introduce \emph{JFSA}, \emph{wFCN}, and the \emph{Segmentation Head}. The Flow Completion Network is pre-trained as a stand-alone module, before plugging it into our pipeline, for which we use the same training recipe as \cite{zhou2023propainter}. We perform the final joint-finetuning of the whole pipeline for an additional 50K iterations.
Additionally, during training we randomly drop the conditioning, to enable classifier-free guidance at inference time. Specifically, we drop the reference image $\refimage$ with a probability of $p_{\text{image}} = 0.15$. At the same time, we drop the flow conditioning and the mask with a probability of $p_{\text{flow}} = 0.1$, $p_{\text{mask}} = 0.1$.

We train the model with a batch size of 16, across 8 NVIDIA A100 GPUs. Training takes approximately 3 days. We keep all the other hyperparameters, including the noise schedule as from Paint-by-Example \cite{yang2023paint}.

\para{Dataset} For training we use YouTube-VOS \cite{xu2018youtube}, a dataset comprising 3471 videos at 6 fps, collected from YouTube. The dataset comes with annotations of the segmentation maps, which we use during training to build our masked source video and the conditioning segmentation map. We resize the frames to a resolution of $448\times256$ and use sequences of $8$ frames to train our model (\ie $T = 8$). The first frame of the sequence serves as the keyframe. For evaluation purposes, we rely on DAVIS \cite{Perazzi2016} and use it to build our evaluation benchmark.

\para{Inference} At inference, we apply the DDIM \cite{ho2020denoising} algorithm with 50 steps for all the visualization in this paper. Moreover, we use the masks predicted by the Segmentation Heads to provide a coherent background in the regions that are not edited. This is achieved with the masked DDIM sampling procedure, as described in \cite{rombach2022high}. 
    
Finally, we enhance the visual quality of the generated results by employing classifier-free guidance \cite{ho2022classifier}. {\methodname} is conditioned on three signals: the driver image, the complete optical flow, and the structure conditioning. Let $\emptyset$ denote the null conditioning signal. Our final prediction can be expressed as:

\begin{equation}
\begin{split}
    \tilde{\epsilon}_\theta(z_t; \refimage, \sourcemask, \sourceflow)  &  = \epsilon_\theta(z_t; \emptyset, \emptyset, \emptyset)+ \\
    + &s_{\text{image}} \cdot \bigl(\epsilon_\theta(z_t; \refimage, \emptyset, \emptyset) -  \epsilon_\theta(z_t; \emptyset, \emptyset, \emptyset)\bigr) \\
    + & s_{\text{mask}} \cdot \bigl(  \epsilon_\theta(z_t; \refimage, \sourcemask, \emptyset) - \epsilon_\theta(z_t; \refimage, \emptyset, \emptyset) \bigr) \\
    +& s_{\text{flow}} \cdot \bigl(  \epsilon_\theta(z_t; \refimage, \sourcemask, \sourceflow) - \epsilon_\theta(z_t; \refimage, \sourcemask, \emptyset) \bigr) \\
\end{split}   
\end{equation}

Throughout this paper, we keep these values fixed to $s_{\text{image}} = 5$, $s_{\text{mask}} = 7$, $s_{\text{flow}} = 6$.

\para{Longer Video Prediction}
Predicting long videos remains an open problem, with many approaches facing challenges in achieving satisfactory results. In this paper, our emphasis is not on improving the length of the videos or enhancing temporal consistency but rather on introducing new tools to enhance editing capabilities. Even though the model is trained for video sequences of length $T = 8$, we generate longer videos using the following approach. We condition the next batch by taking the last frame of the current batch and concatenating it in the temporal dimension to the next batch.  It's important to note that in this case, we don't mask the first frame but provide it clean as input to the 3D Unet. This behavior is replicated during training, where with a probability of $p_{\text{clean}} = 0.1$, we condition input to the model a clean first frame, skipping the JFSA procedure.

\section{Qualitative Results}
\begin{figure*}[p]
    \centering
    \includegraphics{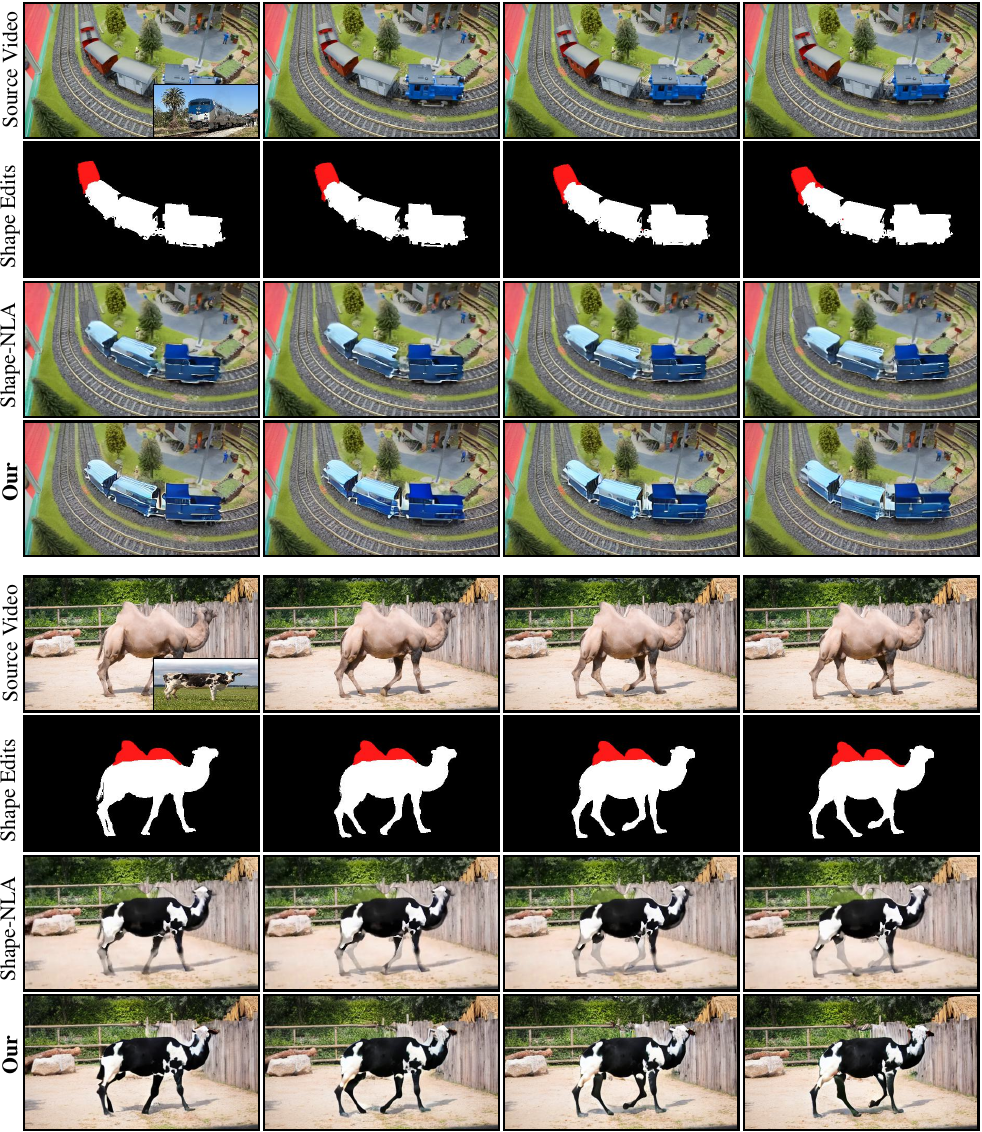}
    \caption{Results for the Joint Appearance-Shape Editing task. Please, note that the shape edit is provided \textbf{only} for the first keyframe, we show all of them only for visualization purposes.}
    \label{fig:supp_shape_editing}
\end{figure*}

\begin{figure*}[p]
    \centering
    \includegraphics{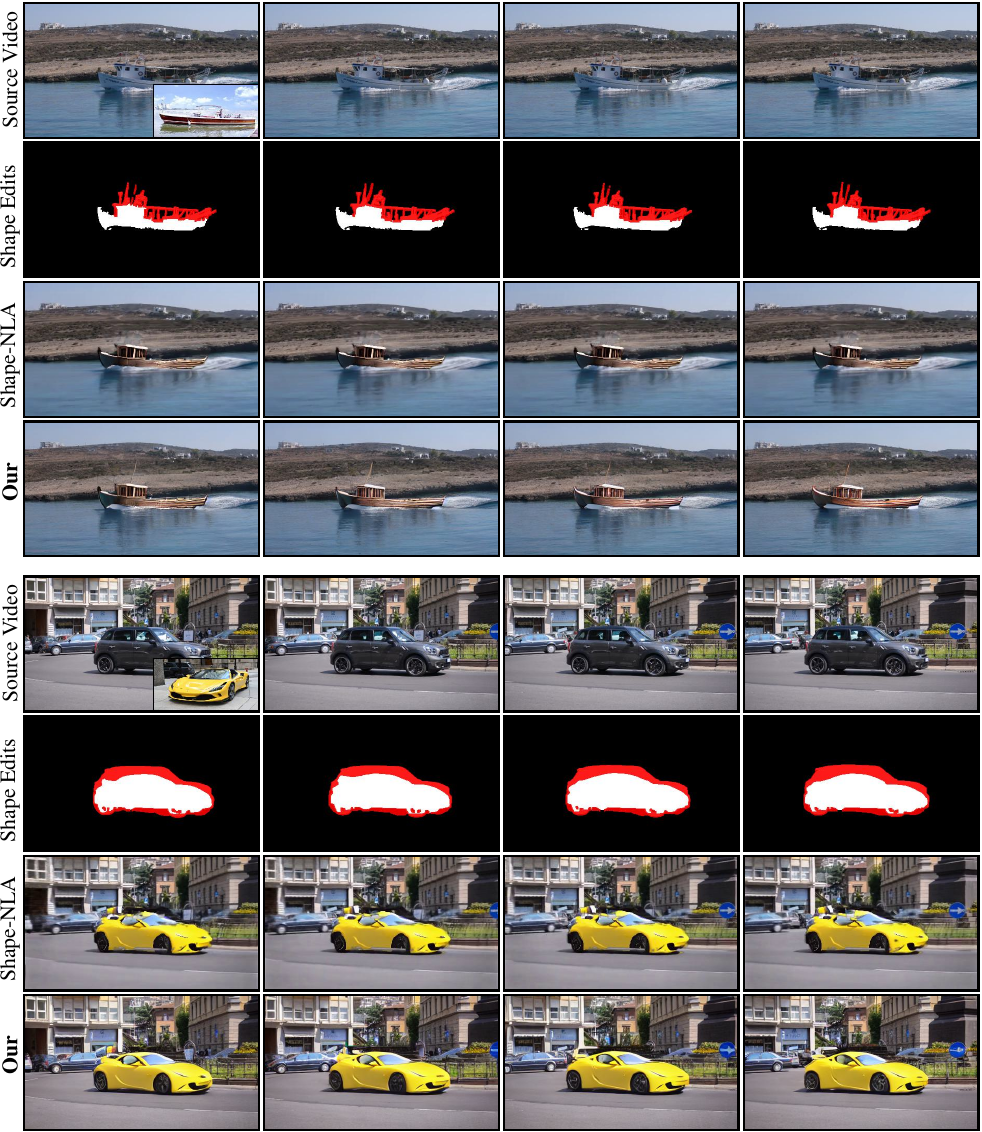}
    \caption{Results for the Joint Appearance-Shape Editing task. Please, note that the shape edit is provided \textbf{only} for the first keyframe, we show all of them only for visualization purposes.}
    \label{fig:supp_shape_editing2}
\end{figure*}

\begin{figure*}[p]
    \centering
    \includegraphics{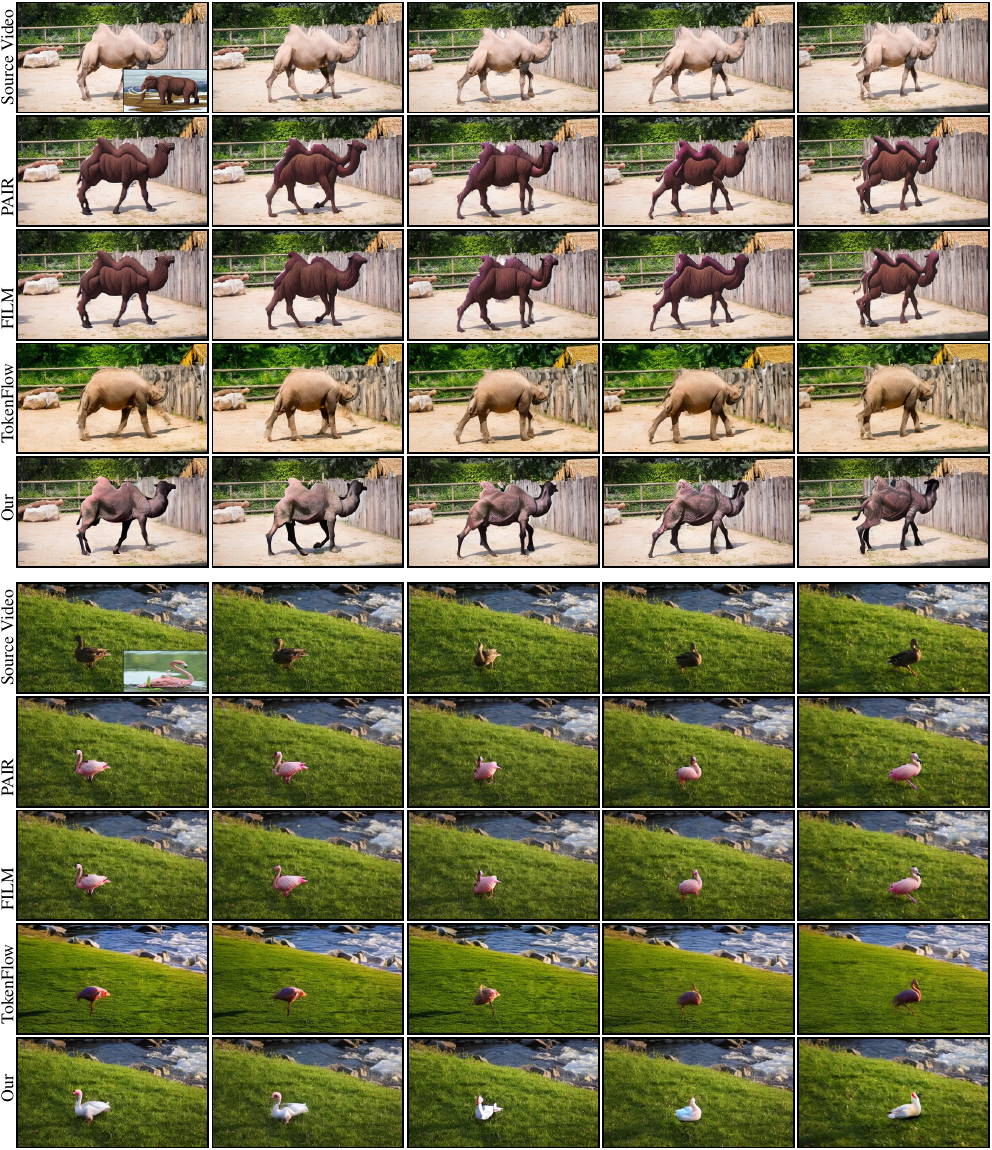}
    \caption{Results for the Image Driven Appearance Editing task. The driver image is displayed in the bottom-left corner of the initial frame in the source video.}
    \label{fig:supp_apperance_editing}
\end{figure*}

\begin{figure*}[p]
    \centering
    \includegraphics{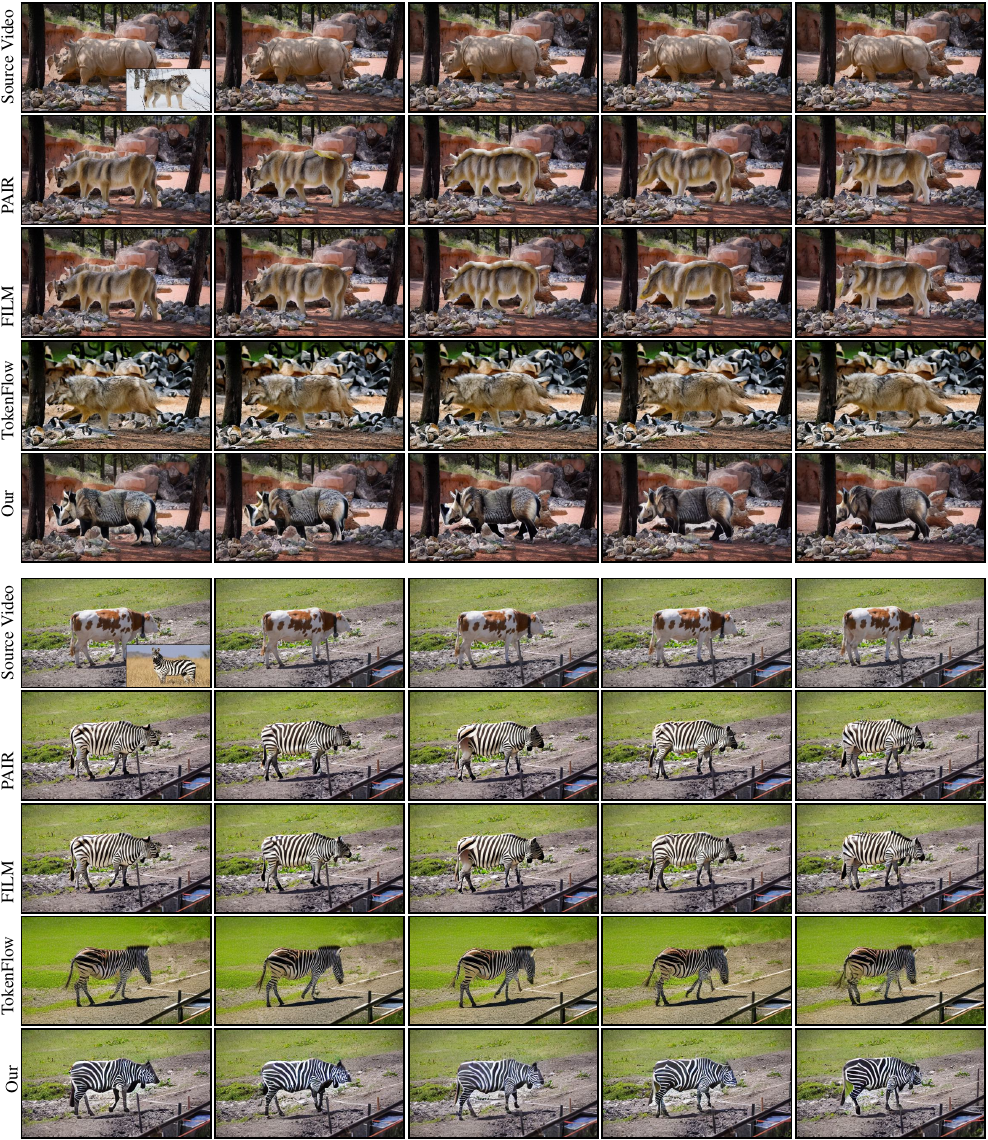}
    \caption{Results for the Image Driven Appearance Editing task. The driver image is displayed in the bottom-left corner of the initial frame in the source video.}
    \label{fig:supp_apperance_editing_2}
\end{figure*}

We report additional qualitative results of {\methodname} and the other methods as shown in the main paper. In \cref{fig:supp_shape_editing} \cref{fig:supp_shape_editing2} we provide additional examples for the joint appearance shape editing, while in \cref{fig:supp_apperance_editing} \cref{fig:supp_apperance_editing_2} we show results for the task of image-driven appearance editing.
\label{sec:supp_qualitatives}


\end{document}